\documentclass{article}
\usepackage[final]{corl_2025}
\usepackage{amssymb}
\usepackage[utf8]{inputenc}
\usepackage{graphicx}
\usepackage{subcaption}
\usepackage{natbib}
\usepackage{amsmath}
\usepackage{hyperref}
\usepackage[normalem]{ulem}
\usepackage[capitalise,noabbrev]{cleveref}
\usepackage{color}
\usepackage{xspace}
\usepackage{enumitem}
\usepackage{comment}
\usepackage{wrapfig}
\usepackage{arydshln}

\def\bm{BusyBox}

\def\pif{$\pi_{0.5}$}
\def\pifft{\pif-canon}
\def\gr{GR00T-N1.6}
\def\grft{\gr-canon}
\def\numdemos{1993}

\def\projecturl{\url{https://microsoft.github.io/BusyBox}}
\def\designfiletype{CAD}

\title{Benchmarking Affordance Generalization with \bm}
\author{
  Dean Fortier\thanks{Microsoft Research,
  \texttt{\{v-defortier,tessh,telascal,michael.murray,akolobov,galenmullins\}\\ 
  @microsoft.com}}\\
  \And
  Timothy Adamson\thanks{Genie, 
\texttt{timadamson21@yahoo.com}. Work done while at Microsoft Research.} \\
  \And
  Tess Hellebrekers\footnotemark[1]\\
  \AND
  Teresa LaScala\footnotemark[1]\\
  \And
  Kofi Ennin\thanks{Mississippi State University, 
\texttt{kofienninacheampong@gmail.com}. Work done while at Microsoft Research.}\\
  \And
  Michael Murray\footnotemark[1]\\
  \AND
  Andrey Kolobov\footnotemark[1]\\
  \And
  Galen Mullins\footnotemark[1]
}

\begin{document}

\maketitle

\begin{abstract}

Vision-Language-Action (VLA) models have been attracting the attention of researchers and practitioners thanks to their promise of \emph{generalization}. Although single-task policies still offer competitive performance~\cite{pmlr-v270-zhao25b}, VLAs are increasingly able to handle commands and environments unseen in their training set~\cite{intelligence2025pi05visionlanguageactionmodelopenworld}. While generalization in vision and language space is undoubtedly important for robust versatile behaviors, a key meta-skill VLAs need to possess is \emph{affordance generalization} -- the ability to manipulate new objects with familiar physical features.

In this work, we present \bm, a physical benchmark for systematic semi-automatic evaluation of VLAs' affordance generalization. \bm\ consists of 6 modules with switches, sliders, wires, buttons, a display, and a dial. The modules can be swapped and rotated to create a multitude of \bm\ variations with different visual appearances but the same set of affordances. We empirically demonstrate that generalization across \bm\ variants is highly challenging even for strong open-weights VLAs such as \pif\ and \gr. To encourage the research community to evaluate their own VLAs on \bm\ and to propose new affordance generalization experiments, we have designed \bm\ to be easy to build in most robotics labs. We release the full set of \designfiletype\ files for 3D-printing its parts as well as a bill of materials for (optionally) assembling its electronics. We also publish a dataset of language-annotated demonstrations that we collected using the common bimanual Mobile Aloha robot~\cite{fu2024mobile} on the canonical \bm\ configuration. All of the released materials are available at \projecturl.
\end{abstract}

\section{Introduction}

\begin{figure*}[t]
    \centering
    \begin{subfigure}[t]{0.31\textwidth}
        \centering
        \includegraphics[width=\linewidth]
        {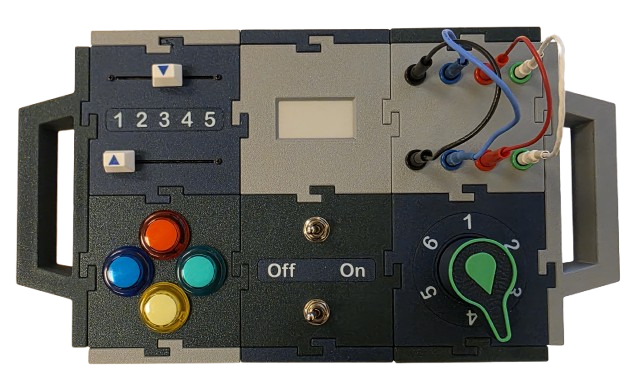}
        \subcaption{canonical}
        \label{fig:canonical}
    \end{subfigure}
    \hfill
    \begin{subfigure}[t]{0.31\textwidth}
        \centering
        \includegraphics[width=\linewidth]
        {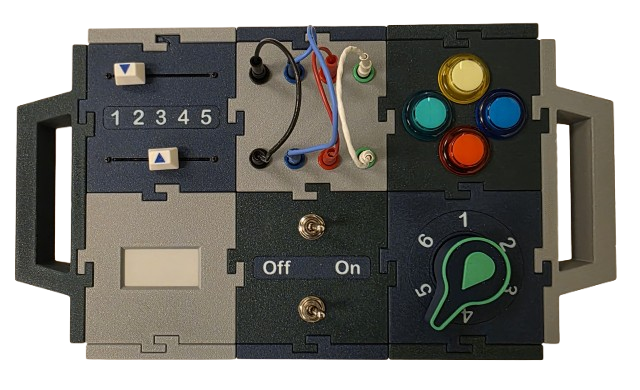}
        \subcaption{semi-shuffled}
        \label{fig:other1}
    \end{subfigure}
    \hfill
    \begin{subfigure}[t]{0.31\textwidth}
        \centering
        \includegraphics[width=\linewidth]
        {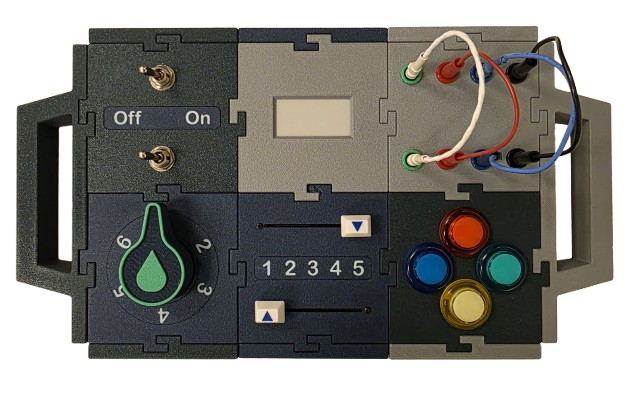}
        \subcaption{fully shuffled}
        \label{fig:other2}
    \end{subfigure}
    \caption{\bm\ configurations used in our experiments. All of them consists of 6 modules: \emph{buttons, display, knob, sliders, switches,} and \emph{wires}. These modules can be swapped with each other and rotated. \textbf{(a)} is the \textbf{canonical} configuration, on which we collected a dataset of demonstrations (see \Cref{fig:data_pie}). \textbf{(b)} is a \textbf{semi-shuffled} configuration: the positions of \emph{buttons, wires,} and \emph{display} modules are different w.r.t. the canonical configuration in (a), and the \emph{buttons} module is also rotated upside down. \textbf{(c)} is \textbf{fully shuffled} -- all 5 manipulable modules are different in position or orientation compared to the canonical \bm: \emph{buttons, knob, sliders,} and \emph{switches} are moved, and the \emph{wires} module is flipped. Many other shuffled \bm\ configurations are possible as well.}
    \label{fig:Taskboxes}
    \vspace{-0.3in}
\end{figure*}

\emph{Vision-Language-Action (VLA)} models~\cite{pmlr-v229-zitkovich23a},  also known as \emph{Robot Foundation Models}~\cite{bommasani2022opportunitiesrisksfoundationmodels}, hold the promise of revolutionizing robot control as much as large language models have advanced natural language processing. In particular, they aim to make robot behaviors \emph{general}, i.e., applicable across robot embodiments, tasks, and environments -- including those not represented in the training data. Generalization takes many forms, and robotics researchers have created a number of benchmarks to evaluate them~\cite{james2019rlbenchrobotlearningbenchmark,mees2022calvin,liu2023libero}. Some types of generalization studied in the context of robotics carry over from language and vision, e.g., characterizing a VLA's ability to comprehend new instructions that reference familiar concepts or to recognize camera images of known objects in new environments. However, robust physical interaction also requires VLAs to exhibit robotics-specific generalization flavors, such as understanding how an unfamiliar object can be manipulated based on its appearance and the robot's experience of handling other objects in the past. We call this capability \emph{affordance generalization}~\cite{ju2024roboabcaffordancegeneralizationcategories}. For people, this capability is crucial for successfully going about their daily lives. Indeed, in environments designed for humans, key user interface elements -- buttons, switches, etc -- are purposefully created to look and function similarly across different objects, in order to facilitate the generalization of basic affordances for the average person. 
\emph{We posit that robots need to be able to use basic affordances not only to operate fluently in human environments but also to be \emph{perceived} as intelligent and reliable.}

\begin{wrapfigure}{r}{0.55\textwidth}
    \centering
    \vspace{-0.2in}
    \includegraphics[width=1.0\linewidth]{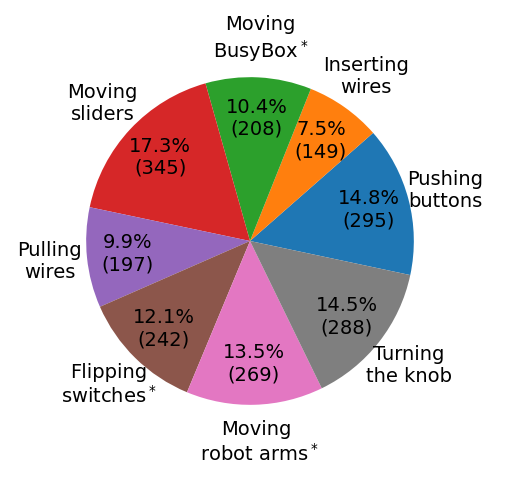}
    \caption{Breakdown of the dataset of 1993 \bm\ demonstrations at \projecturl\ by affordance category. \textbf{* denotes bimanual affordances}, although virtually \bm\ tasks benefit from positioning both robot arms so that their wrist cameras observe the manipulated object up-close.}
    \label{fig:data_pie}
\end{wrapfigure}

This work introduces \bm\ (\Cref{fig:Taskboxes}), an open-source physical benchmark for systematically evaluating basic affordance generalization in VLAs. \bm\ is a 3D-printable device consisting of 6 interlocked modules that have buttons of various colors, switches, sliders with marked positions, wires with pluggable connectors, a knob, and a display. These or similar elements are commonly found on everyday items in home and industrial environments. However, overall, \bm\  looks different from any objects likely to be present in VLAs' training data. A key feature of \bm\ is that its 6 modules can be easily attached in different positions and orientations w.r.t. each other (\Cref{fig:disassembled}), giving rise to a family of distinct \bm\ instances with the same set of basic affordances. A person seeing a \bm\ instance for the first time can learn all its affordances in under a minute, including pulling out and inserting wires, rotating the knob, and reading information on the display. Having learned them on one \bm, the person will be able to use these affordances on any other \bm\ configuration zero-shot. 

How close do VLAs come to this level of affordance generalization? A second contribution of our work is an experiment protocol and a dataset of demonstrations (\Cref{fig:data_pie}) for empirically answering questions like this. We collected \numdemos\ trajectories across 8
task families representing \bm's affordances, e.g., \textit{"rotate the knob to position 4"}, by teleoperating a dual-arm Trossen Mobile Aloha robot~\cite{fu2024mobile} on a designated canonical \bm\ instance (\Cref{fig:canonical}). Next, we used this dataset to finetune \pif~\cite{pi05} and \gr~\cite{gr}, some of the strongest open-weight VLAs, and evaluated the success rates of the resulting  models on a fixed set of task instructions on three \bm\ configurations in \Cref{fig:Taskboxes}: the canonical one, on which the finetuning dataset was collected in the first place; a semi-shuffled one, which differs from the canonical \bm\ in the location of 3 modules; and a fully shuffled one. Our first result (\Cref{fig:res_canon}) shows that the finetuned \pif\ and \gr\ perform reasonably well on the canonical \bm -- the \bm\ configuration on which we collected the training dataset --  indicating that the dataset demonstration quality and diversity is sufficient for learning in-distribution versions of \bm\ tasks. At the same time, evaluation on the semi-shuffled \bm\ configuration revealed (\Cref{fig:res_semi}) that \pif\ and \gr\ finetuned on the data from the canonical \bm\ have difficulties on the visually out-of-distribution \bm\ tasks, especially on the tasks that involve the shuffled modules. On the fully shuffled configuration, \pif's and \gr's performance degraded even more severely (\Cref{fig:res_fully}), despite the fact that both semi- and fully shuffled \bm es have the same set of affordances as the canonical one.

 As these findings demonstrate, basic affordance generalization is a major area for improvement even for the strongest existing VLAs, and \bm\ as a benchmark is far from saturation. Our study is just one example of experiments enabled by \bm. E.g., although we don't explore VLAs' spatial reasoning in this work, \bm\ is highly suitable for evaluating it, on tasks such as \textit{"pull the 2nd wire from the left"}. \bm\ can also be used for assessing the effectiveness of verbal corrections during task execution~\cite{liu2023interactiverobotlearningverbal,shi2024yellrobot}. More broadly, \bm\ was inspired by the interactive game \textit{Keep Talking and Nobody Explodes}~\cite{pestaluky2015keeptalking}, which involves two players who need to communicate with each other in order to determine and execute a sequence of actions that defuses a bomb that \bm\ represents. As an implementation of this game, \bm\ can be a rich environment for evaluating physical and digital AI agents and studying human-robot interaction. 


In summary, our work's contributions are:
\begin{itemize}[leftmargin=*]
    \item \bm, a physical benchmark for evaluating basic affordance generalization and \designfiletype\ designs for 3D-printing it. 
    
    \item A language-annotated dataset of \numdemos\ manipulation demonstrations, many of them bimanual, gathered on \bm's affordances.
    \item An experiment protocol for using \bm\ and this dataset or its equivalents for assesssing affordance generalization in VLAs, along with baseline empirical results on \pif\ and \gr.
\end{itemize}
All of the released materials are available at \projecturl.
\section{\bm\ design and instrumentation}

\bm\ follows a modular architecture that uses easily reproducible 3D-printed components and instrumentation. All CAD files, source code, and instructions referenced in this section are available on the project page at \projecturl.

\subsection{Modules}

\bm\ comprises six distinct modules shown in \Cref{fig:canonical}, each representing a basic affordance. These modules are visually intuitive and engineered to be well within the physical manipulation capabilities of off-the-shelf 6-DoF robot arms with two-finger grippers:

\begin{itemize}[leftmargin=*]
    \item \textbf{Display module.} This module features an E Ink display and three LED indicators. It houses the main electronics of the \bm\ and provides visual feedback to both the VLA and the user. The E Ink display enhances information visibility for robots' cameras.
    \item \textbf{Buttons module.} Consisting of four colored, illuminated buttons, this module requires the robot to press the correct button using a single arm. Rotating the module serves as a method to test policy generalization, as agents may otherwise memorize button positions.
    \item \textbf{Sliders module.} This module includes two sliders, each adjustable between values 1 and 5. The primary challenge lies in  determining which slider to move and ensuring accurate positioning.
    \item \textbf{Knob module.} The knob can be rotated to a specified value between 1 and 6 and has a handle to facilitate its manipulation. 
    Occlusions can complicate rotating the knob to a desired position.
    \item \textbf{Switches module.} Featuring switches with on/off positions, this module is visually simple but presents learning challenges: due to the force required to flip the switch, one robot arm needs manipulate the switch while the other needs to pin \bm\ in place, making this a bimanual affordance.
    \item \textbf{Wire Module.} This module involves inserting or unplugging colored wires. 
\end{itemize}

\subsection{3D-printable design}

\begin{wrapfigure}{r}{0.50\textwidth}
    \centering
    \vspace{-0.2in}
    \includegraphics[width=1.0\linewidth]
    {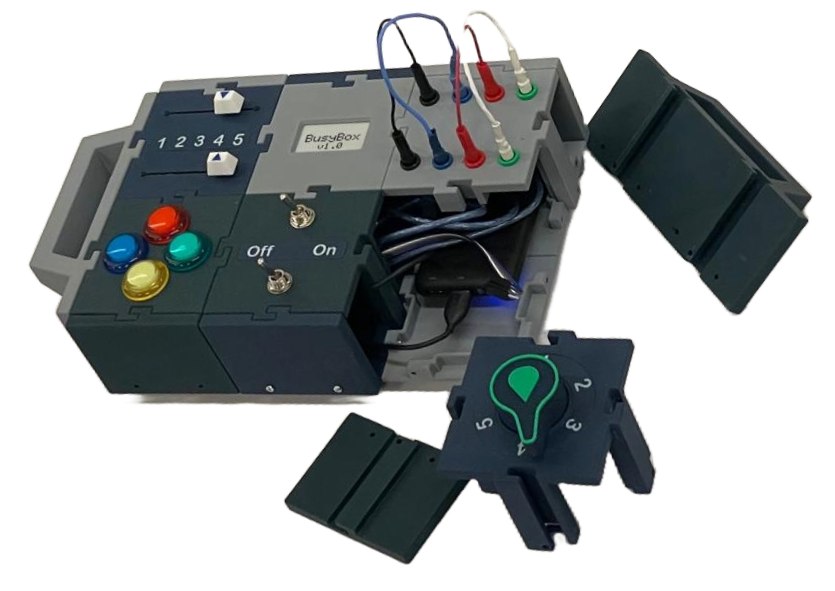}
    \caption{Disassembled \bm\ }
    \label{fig:disassembled}
    \vspace{-0.2in}
\end{wrapfigure}

One of our principal objectives in developing the \bm\ was to facilitate its replication in research laboratories. \bm\ also needed to be robust enough to survive rough handling as well as light enough to be picked up and moved by standard bimanual manipulators used in robot learning research. E.g., widespread Trossen Aloha systems with ViperX follower arms can lift only a 750 gram payload per arm. To this end, the body of \bm\ employs exclusively 3D-printed components in order to maximize \bm's strength-to-weight ratio and make it easy to manufacture.

As shown in \Cref{fig:disassembled}, \bm's components interlock using snap connectors, which streamlines both assembly and disassembly, thereby enabling rapid \bm\ reconfiguration. The perimeter of the \bm's interlocked modules is covered by side elements. Two of the side elements have handles, adding the affordances of conveniently picking up and rotating the \bm. The canonical \bm\ instance (\Cref{fig:canonical}) has 6 distinct modules in specific orientations, arranged in a three-by-two grid. To obtain other configurations, such as those in \Cref{fig:other1} and Figure 1c, the modules can be easily permuted and rotated with respect to each other by 90, 180, or 270 degrees. \bm's design also allows  multiple instances of the same module, alternative module arrangements such as two-by-two or two-by-one, and assemblies involving more than 6 modules.

 We produced our \bm\ using a two-filament printer capable of printing in different colors. This enhances the legibility of textual elements, as demonstrated with high-contrast color combinations seen in \Cref{fig:disassembled}. For users lacking access to two-filament color printers, we have verified that single-color prints with manually painted highlights including the position numbers for the sliders and the knob are a practical alternative.

\subsection{Instrumentation}

To make both policy learning and evaluation easier, it is useful to have a mechanism that automatically records the state of all of \bm's controls at every time step. This not only helps determine if a task has been accomplished but is also valuable for tracking \emph{progress}~\cite{black2024pi0visionlanguageactionflowmodel} when a task consists of several substeps, e.g., \emph{"move the top slider to 2, insert the red wire, and press the leftmost button"}. With this in mind, we provide readily replicable \emph{optional} electronic instrumentation that registers \bm\ state for live monitoring and/or recording it into demonstration trajectory files. We emphasize that even without this instrumentation, the \bm\ can be used for VLA evaluation.  

The central control unit is a Raspberry Pi 0, which resides in the primary module alongside an E Ink display. This module must be included in the \bm\ configuration for the instrumentation to be functional. The Raspberry Pi 0 is augmented with a USB expansion board, enabling all individual components to interface via USB connections. We specifically selected the electrical components to be widely available and pre-instrumented to function as USB devices via Arduino serial communication. This architecture supports plug-and-play compatibility between \bm\ elements and the display module, enabling users to interchange modules freely. The instrumentation measurements are broadcast on the network at 10 Hz using a Raspberry Pi.

\begin{wrapfigure}{r}{0.5\textwidth}
    \centering
    \includegraphics[width=1.0\linewidth]
    {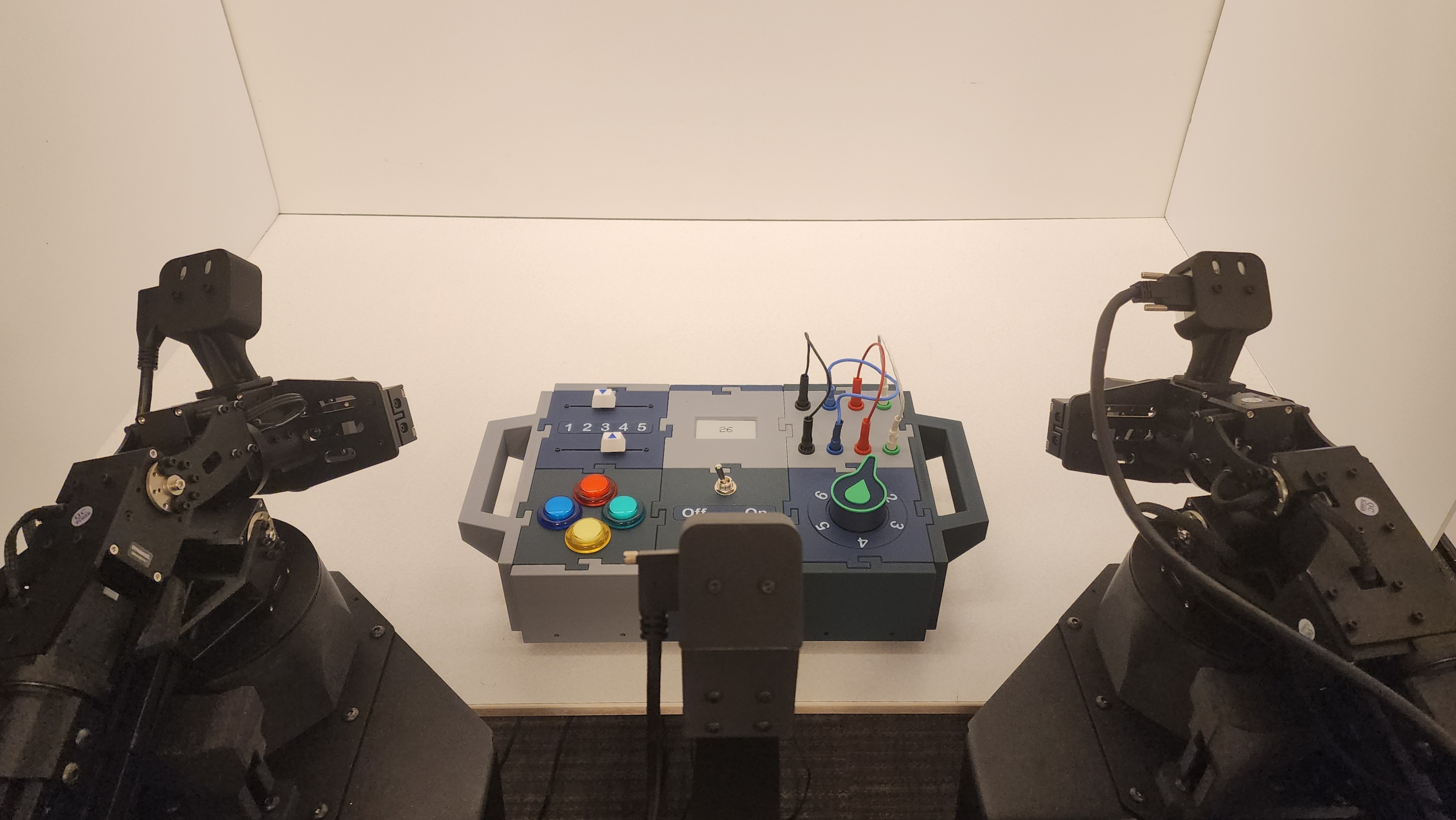}
    \caption{Illustration of our \bm\ data collection setup based on Mobile Aloha.} 
    \label{fig:data_collection_fixture}
    \vspace{-0.2in}
\end{wrapfigure}

The wireless networking capabilities of the Raspberry Pi 0 allow users to remotely access and monitor the state of the \bm. When first powered on, the primary module displays the connection information on the E Ink display. For greater control, the users may also connect to the primary module directly via a USB-C interface. Power is supplied to all modules through the Raspberry Pi 0’s USB ports, with the option of either a wired connection or a battery-powered setup. For battery operation, a standard 2500 mAh power brick is housed within the \bm.
\section{Data collection \label{sec:datacoll}}

To illustrate affordance generalization experiments enabled by \bm, we collected a finetuning dataset for adapting VLAs on \bm's affordances (\Cref{fig:data_pie}, \Cref{tab:task_types}). In this section we detail our data collection protocol and the dataset itself. 

\subsection{\bm\ dataset}

We collected \numdemos\ demonstrations data for the \bm\ task instructions listed in \Cref{tab:task_types}, with the breakdown of the number of demonstrations across affordance categories listed in \Cref{fig:data_pie}. Each affordance category was represented by by several possible language instructions. The task instructions referenced color and/or desired state of the manipulated controls or of \bm\ itself. Some of the instructions also specified whether to use the right or left gripper. Our data collection workspace is shown in \Cref{fig:data_collection_fixture}.

\begin{table*}[t]
    \centering
    \small
    \begin{tabular}{p{0.22\linewidth} p{0.77\linewidth}}
        \hline
        \textbf{Affordance category} & \textbf{Task instruction variants} \\
        \hline
        Pushing buttons 
            & ``Push the \{color\} button with the \{left, right\} gripper.'' \\
        Moving sliders 
            & ``Move the \{top, bottom\} slider to position\{1, 2, 3, 4, 5\}.'' \\
        Turning the knob 
            & ``Turn the knob to position \{1, 2, 3, 4, 5, 6\}.'' \\
        Flipping switches 
            & ``Flip the \{left, right\} switch \{on, off\} with the \{left, right\} gripper.'' \\
        Pulling wires 
            & ``Pull the \{red, black, blue, white\} wire.'' \\
        Inserting wires 
            & ``Insert the \{red, black, blue, white\} wire.'' \\
        Moving \bm\ 
            & ``Rotate \bm\ \{clockwise, counter-clockwise\}'' \\
            & ``Move \bm\ \{left, right, closer, away\}'' \\
        Moving robot arms
            & ``View \bm\ from above.'' \\
            & ``Move \{left, right\}, gripper to the \{left, right\}.'' \\
            & ``Open both grippers.'' \\
            & ``Open \{left, right\} gripper.'' \\
        \hline
    \end{tabular}
    \caption{Task/affordance types and instruction variants.}
    \label{tab:task_types}
    \vspace{-0.2in}
\end{table*}

\subsection{Teleoperation instructions}




\subsubsection{Randomizing the initial state \label{sec:init_state}}

Ensuring the diversity of initial states of the demonstrations is crucial for state coverage in the data and for learning robust policies. Given the number of factors of variation in the environment (positions of the sliders, the switches, the \bm\ itself, the robot, etc), we chose not to rely on the teleoperator to randomize the initial states along all these dimensions. Instead, generation of initial states was directed by a script that gave instructions to the teleoperator before the start of each demonstration. The parameters of each instruction were sampled uniformly at random, unless stated otherwise. Before the teleoperator executed these instructions, the teleoperator would verify that the robot and the \bm\ pose configuration and pose were as in \Cref{fig:data_collection_fixture}: the robot faced the \bm\ directly, the \bm\ wasn't rotated, the \bm\ was centered along the near edge of the workspace 2 inches from that edge, and all the wires were inserted.

\begin{itemize}[leftmargin=*]
    \item \emph{Set the top slider to position $s_{top}$ and the bottom slider to $s_{bottom}$}, where $s_{top}, s_{bottom} \sim \{1,2,3,4,5, \mbox{between 1 and 2}, \ldots, \mbox{between 4 and 5} \}$.

    \item \emph{Set the knob to position $s_{knob}$}, where $s_{knob} \sim \{1,2,3,4,5,6, \mbox{between 1 and 2}, \ldots, \mbox{between 6 and 1} \}$.

    \item \emph{Set the top switch to position $sw_{top}$ and the bottom switch to $sw_{bottom}$}, where $sw_{top}, sw_{bottom} \sim \{\mbox{on}, \mbox{off}\}$.

    \item With $p = 0.5$, leave all the wires inserted. Otherwise, for each wire independently with $p = 0.5$, pull out one end of that wire. 

    \item \emph{Rotate the \bm\ by roughly $\phi$ degrees clockwise}, where $\phi \sim \{-20, -10, 0, 10, 20\}$.

    \item \emph{Move the \bm\ roughly $x$ inches away and $y$ inches to the right}, where $x \sim \{0, 2, 4\}$, $y \sim \{-4, -2, 0, 2, 4\}$.

    \item \emph{Rotate the robot by roughly $\theta$ degrees clockwise}, where $\theta \sim \{-10, 10\}$.
    
\end{itemize}

The initial state for each demonstration episode was sampled in a way that guaranteed that the goal described in the task instruction sampled for that episode wasn't achieved in that initial state. E.g., if the task instruction was \emph{``Turn the knob to position 4 ''}, then in the initial state sampling process, '4' was excluded from the set of possible values for the knob.

\subsubsection{Teleoperation style}

To mitigate bias and unpredictable behavior in the finetuned models, the teleoperators were asked to follow several guidelines:

\begin{enumerate}[leftmargin=*]
    \item Be efficient with movement. Demonstrations for tasks other than wire insertion should take no longer than 20 seconds. For wire insertion, demonstrations should not exceed 45 seconds.
    \item Keep moving. Demonstrations should be active, only remain still when something dynamic is happening (i.e. dropping a wire).
    \item End the demonstration when the requested task has been completed. There is no need to return to a neutral position.
    \item Whenever possible, ensure that both wrist cameras have a view of the parts of the environment relevant for task at hand.
\end{enumerate}

In most data collection episodes, the robot arms started at the same ``home" position. However, we also collected several hundred recovery demonstrations, where the teleoperators were asked to begin the episode with robot arms at one of the poses where VLA policies learned without these corrective demonstrations tended to get stuck. These extra demonstrations proved especially helpful for learning \emph{pulling wires} and \emph{turning the knob} affordances.

Additionally, teleoperators had to follow affordance-specific instructions:

\begin{enumerate}[leftmargin=*]
    \item \textbf{Buttons}: With grippers closed, press the button with one gripper. End recording just after the button is released and the gripper is a few inches above the button. 
    \item \textbf{Sliders}: With grippers closed, move the slider by pushing it in one direction or another until the slider reaches the position mentioned in the task description. If the target value is overshot, correct the overshoot and end the recording.
    \item \textbf{Pulling Wires}: Pull out only the end of the wire closer to the robot, leaving the other end inserted. 
    \item \textbf{Inserting Wires}: When inserting a wire, the demonstration should not exceed 45 seconds.
    \item \textbf{Switches}: Brace the box by inserting one gripper into one handle and move the switch to the desired position with the other gripper.
    \item \textbf{Knobs}: With grippers closed, one gripper should nudge the knob until it is in the desired position while the other should view the knob with its camera. If the target position is overshot, nudge the knob in the other direction. If the box moves while turning the knob, brace the \bm\ with the other arm.
\end{enumerate}

\section{Affordance generalization experiment}

Using the dataset described in \Cref{sec:datacoll}, we conducted an experiment on a real Trossen Mobile Aloha robot, meant to illustrate the utility of \bm\ by answering the conceptual question: \textbf{how well do strong open-weights VLAs do on affordance generalization?} In particular, we asked: \textbf{how well do strong VLAs finetuned on data collected on the canonical \bm\ (\Cref{fig:canonical}) perform on the semi- and fully shuffled \bm\ variants in \Cref{fig:other1} and \Cref{fig:other2}}?

\subsection{Experiment setup}

\textbf{Models.} We used the official versions of \pif~\cite{pi05} and \gr~\cite{gr}, available at  \url{https://github.com/Physical-Intelligence/openpi} and \url{https://github.com/NVIDIA/Isaac-GR00T}, respectively,  as the base VLAs. We finetuned each of them on the entire dataset from the canonical \bm\ configuration (\Cref{fig:data_pie}). The finetuned versions are denoted as \pifft\ and \grft.\\

\textbf{Evaluation protocol.} We sampled a set of 60 task instructions from \Cref{tab:task_types}, with 10 task instructions from \emph{moving \bm}, \emph{moving sliders}, \emph{pulling wires}, \emph{pushing buttons}, \emph{turning the knob}, and \emph{flipping switches} affordance categories, and evaluated \pifft\ and \grft\ on each of the three \bm\ configurations in \Cref{fig:Taskboxes} on this instruction set. The instructions were given to the models one-at-a-time, and the initial state the \bm\ and the robot was reset before each instruction as described below. We omitted the \emph{inserting wires} category, because neither \pifft\ nor \grft\  was able to consistently achieve a 10\% success rate on it even on the canonical \bm\ configuration. We also omitted the \emph{moving the robot arms} tasks such as \emph{``close both grippers''}, because they were very easy for both \pifft\ and \grft, and their execution doesn't require paying attention to the \bm. Within each category, task instructions differed in the object they referred to and, in some cases, the arm with which the task was to be performed (see \cref{tab:task_types}). For example, the \emph{push button} category included instructions such as \emph{``push the green button with the left gripper"} and \emph{``push the blue button with the right gripper"}. Across all the three \bm\ configurations in our experiment, the 60 task instructions were presented to the models in the same order.

For each \bm\ configuration, the model evaluation was done in \emph{randomly-ordered paired episodes}. Namely, for each \bm\ configuration, for each of the 60 instructions, our script chose the initial \bm\ state (the state of all of its controls and its orientation). Both models were then evaluated on this instruction starting from this \bm\ state in an order that was sampled randomly and independently for each instruction, one model right after another, with a human operator resetting \bm\ to the prescribed initial state inbetween. The robot would start both models' episodes from the same home position. This protocol minimized potential environment drift that may have otherwise affected each model's evaluation conditions. \\

\textbf{Success criteria.} For each instruction, each model's evaluation episode lasted 30s. The episode was considered successful if both of the following conditions were met:

\begin{itemize}[leftmargin=*]
\item The task was complete at the 30s-mark.
\item At the 30s-mark, the \bm\ controls not mentioned in the task instruction remained as in the initial state.
\end{itemize}

For example, if the robot was asked to \emph{``move the top slider to position 4"} and did so while also accidentally rotating the knob from 2 to 1 and leaving the knob at 1 at the 30th second, the episode counted as a failure. Similarly, if, given the same instruction, the robot moved the top slider to 4 and then proceeded to move it to 5, leaving the slider at 5 until the 30th second, that episode counted as a failure as well. 

\subsection{Results}

\begin{figure}[t]
    \centering

    \begin{subfigure}[b]{0.5\textwidth}
        \centering
        \includegraphics[width=\linewidth]{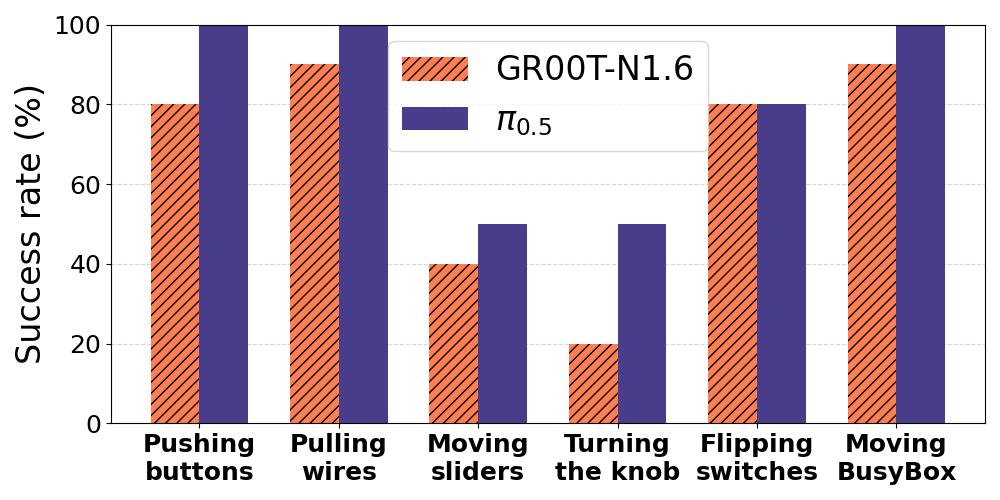}
        \subcaption{Perf. on the canonical \bm\ (\Cref{fig:canonical})}
        \label{fig:res_canon}
    \end{subfigure}

    \vspace{0.3cm}

    \begin{subfigure}[b]{0.49\textwidth}
        \centering
        \includegraphics[width=\linewidth]{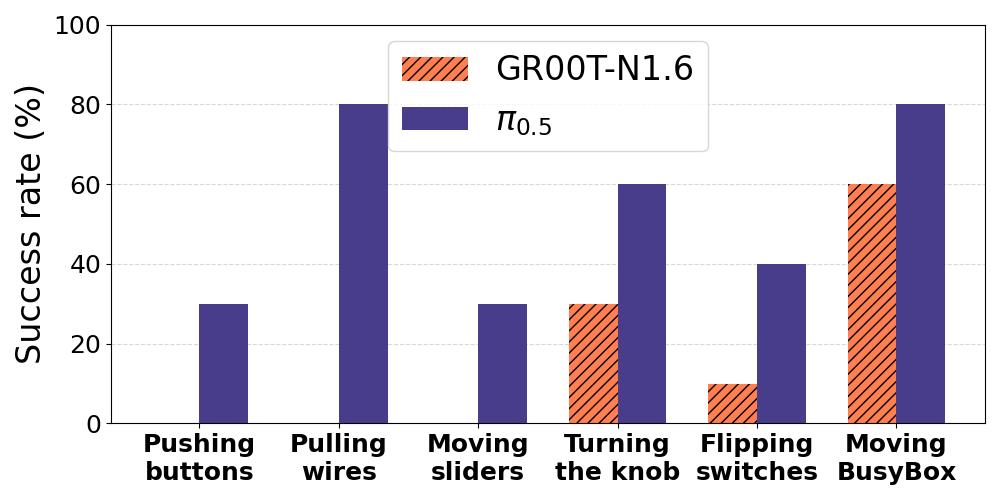}
        \subcaption{Perf. on the semi-shuffled \bm\ (\Cref{fig:other1})}
        \label{fig:res_semi}
    \end{subfigure}
    \hfill
    \begin{subfigure}[b]{0.49\textwidth}
        \centering
        \includegraphics[width=\linewidth]{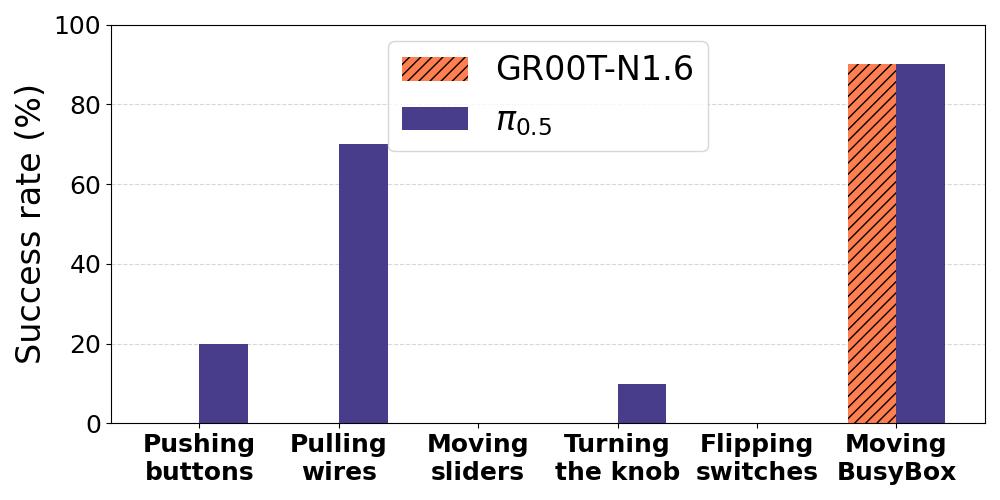}
        \subcaption{Perf. on the fully shuffled \bm\ (\Cref{fig:other2})}
        \label{fig:res_fully}
    \end{subfigure}

    \caption{Results of the affordance generalization experiment: despite all \bm\ configurations being affordance-wise in-distribution w.r.t. the training dataset (\Cref{fig:data_pie}), \pifft\ and \grft\ performed well only on the visually in-distribution canonical \bm\ configuration (\Cref{fig:res_canon}), struggling with the visually out-of-distribution shuffled variants (\Cref{fig:res_semi} and \Cref{fig:res_fully}).}
    \label{fig:results}
\end{figure}

The results are presented in \Cref{fig:results}. When analyzing them, it is useful to view \Cref{fig:res_canon} as showing performance on a \bm\ configuration that is \textbf{visually in-distribution (VID)} for \pifft\ and \grft, and \Cref{fig:res_semi} and \Cref{fig:res_fully} as describing performance on the \textbf{visually out-of-distribution (VOOD)} \bm\ variants, while all three variants are \textbf{affordance-wise in-distribution (AID)}.\\

\noindent
\textbf{Notable behavioral patterns}. According to \Cref{fig:res_canon}, both \pifft\ and \grft\ performed fairly well in the VID/AID setting, showing that the quality and diversity of our \bm\ dataset (\Cref{fig:data_pie}) is sufficient for learning \bm\ affordances mentioned in this  plot via supervised VLA finetuning.

Qualitatively, however, \pifft's and \grft's behaviors had marked differences:
\begin{itemize}[leftmargin=*]
    \item \textbf{Behavior after goal attainment.} Recall that during evaluation each model's episodes were terminated only after 30s, even if the model completed the task early. Once a task was completed, \pifft\ would normally stop and do nothing until the end of the episode. \grft\ would instead often continue to adjust \bm's controls even if it was unnecessary, sometimes ``unsolving" the task.
    \item \textbf{VOOD behavior.} When a task instruction required dealing with a VOOD-positioned \bm\ module for the experiments in \Cref{fig:res_semi} and \Cref{fig:res_fully}, e.g., pushing a button on the semi-shuffled \bm, \pifft\ would often freeze until the end of the episode. \grft\ would instead almost always continue to move the arms to the location where this module \emph{would be positioned on the canonical \bm}. This hints at \grft\ implicitly memorizing module positions from the training data by overfitting to proprioception, and paying insufficient attention to Mobile Aloha's scene camera, which had all of \bm's modules in view at the start of each episode, independently of the specific \bm\ configuration. Despite these differences, the effect of these behaviors in \pifft and \grft\ was the same, leading to task failures and accounting for most of the failed episodes in \Cref{fig:other1} and \Cref{fig:other2}. 
\end{itemize}

\noindent
\textbf{VID/AID performance (canonical \bm, \Cref{fig:res_canon}).} In the visually in-distribution setting, the models performed comparably. Both consistently tried to reach for the \bm\ module relevant to the task instruction. The \grft's slightly lower success rate was due to its motions occasionally being less precise and its aforementioned tendency to disturb the \bm\ even after successful task completion, sometimes destroying the goal state. \\

\noindent
\textbf{VOOD/AID performance (semi-shuffled \bm, \Cref{fig:res_semi}).} Compared to \Cref{fig:res_canon}, the success rates of both models significantly suffered from the module rearrangement. This may seem expected for the affordances involving the shifted modules -- \emph{buttons} and \emph{wires}. However, the performance was, in fact, affected across all affordance categories, e.g., \emph{flipping switches}. We hypothesize that this is due to the visual appearance of every module's \emph{neighborhood} changing, even if the module itself stayed in place (compare \Cref{fig:canonical} and \Cref{fig:other1}). For instance, although the \emph{switches} module remained as in the canonical \bm\ configuration, the presence of the \emph{wires} module right above it on the semi-shuffled \bm\ likely confused the models. \\

\noindent
\textbf{VOOD/AID performance (fully shuffled \bm, \Cref{fig:res_fully}).} Given the results on the semi-shuffled \bm, the models' poor performance on the fully shuffled \bm\ is unsurprising. Note that the \emph{wires} module was merely flipped, not moved, but this alone confused \grft\ sufficiently to fail every attempt at wire manipulation. Many of \grft's attempts actually succeeded in pulling out a wire, but invariably of the wrong color. \pifft\ tended to discern wire colors better once its wrist cameras got close to the \emph{wires} module, but \pifft's success rate on \emph{pulling wires} suffered nonetheless. For both models, the most successful affordance category was \emph{moving \bm}, since it only weakly depends on the positions of individual modules. \\

In summary, our experiment shows that affordance generalization is a challenge even for some of the strongest existing VLAs and even in in-distribution settings, because it can require out-of-distribution generalization in the visual space. \bm\ can be a helpful tool in evaluating and improving this aspect of VLAs' performance.

\section{Related Work}

Compared to the existing benchmarking methods for robot manipulation models, the novelty of our work is in introducing a physical benchmark for systematic evaluation of affordance generalization.

\subsection{Physical Benchmarks}

Physical benchmarks are common in robotic manipulation research. Examples include the Functional Manipulation Benchmark (FMB)~\cite{luo2024fmbfunctionalmanipulationbenchmark}, FurnitureBench~\cite{heo2023furniturebench}, NIST Task Boards~\cite{nist2020}, and Digital Robot Judge Task Boards (DR. J.) ~\cite{judge2024}. Like \bm,  NIST Task Boards and FurnitureBench involve manipulating deformable and articulated objects such as wires and switches. Their focus, however, is on assessing a model's ability to drive difficult contact-rich manipulation rather than its ability to generalize. FMB has been used to benchmark generalization, but not specifically in the affordance space. DR. J. logs changes in the box state, which can be used to verify task completion but, like FMB, doesn't evaluate affordance generalization.

\subsection{Simulated Benchmarks}

Simulation-based benchmarks provide consistent and controllable environments that make task completion easier to measure reliably and record automatically. However, for modeling objects such as switches, wires, and sliders, with all their imperfections that frequently hamper manipulation, simulation has so far been too imprecise, too slow, or both. Common simulated task suites, including LIBERO~\cite{liu2023libero} and SimplerEnv~\cite{li24simpler}, don't involve complex objects like these, and evaluating affordance generalization isn't their focus. The simulation-based benchmark that involves operations most similar to those on \bm\ is BusyBoard~\cite{liu2022busybotlearninginteractreason}. However, BusyBoard is still very distinct, lacks open-source CAD files for reproducing it, and, like the other benchmarks, sidesteps evaluating affordance generalization. 
\section{Conclusion}

We presented \bm - a modular, reconfigurable \bm\ for benchmarking affordance generalization in VLAs on physical hardware. \bm\ targets basic affordances such as flipping switches and plugging in audio cables, which are ubiquitous in home and industrial environments but are underrepresented in existing datasets and benchmarks. \bm\ comes with open-source CAD files and electronics design for easy reproduction in robotics research labs. Using a demonstration dataset, which we collected on \bm\ and are also open-sourcing with this work, we conducted an illustrative empirical study showing that strong VLAs such as \pif\ and \gr\ currently struggle with affordance generalization. We hope that \bm\ will encourage more active and more reproducible research on this subject.

\bibliography{references}

\end{document}